# Multiclass Segmentation using Teeth Attention Modules for Dental X-ray Images

Afnan Ghafoor[1], Seong-Yong Moon[2] and Bumshik Lee[1], Member, IEEE
[1]Department of Information and Communication Engineering, Chosun University
[2]Department of Oral and Maxillofacial Surgery, College of Dentistry, Chosun University

Corresponding author: Bumshik Lee (e-mail: bslee@chosun.ac.kr).

**ABSTRACT** This paper proposed a cutting-edge multiclass teeth segmentation architecture that integrates an M-Net-like structure with Swin Transformers and a novel component named Teeth Attention Block (TAB). Existing teeth image segmentation methods have issues with less accurate and unreliable segmentation outcomes due to the complex and varying morphology of teeth, although teeth segmentation in dental panoramic images is essential for dental disease diagnosis. We propose a novel teeth segmentation model incorporating an M-Net-like structure with Swin Transformers and TAB. The proposed TAB utilizes a unique attention mechanism that focuses specifically on the complex structures of teeth. The attention mechanism in TAB precisely highlights key elements of teeth features in panoramic images, resulting in more accurate segmentation outcomes. The proposed architecture effectively captures local and global contextual information, accurately defining each tooth and its surrounding structures. Furthermore, we employ a multiscale supervision strategy, which leverages the left and right legs of the U-Net structure, boosting the performance of the segmentation with enhanced feature representation. The squared Dice loss is utilized to tackle the class imbalance issue, ensuring accurate segmentation across all classes. The proposed method was validated on a panoramic teeth X-ray dataset, which was taken in a real-world dental diagnosis. The experimental results demonstrate the efficacy of our proposed architecture for tooth segmentation on multiple benchmark dental image datasets, outperforming existing state-of-the-art methods in objective metrics and visual examinations. This study has the potential to significantly enhance dental image analysis and contribute to advances in dental applications.

## I. INTRODUCTION

Dental imaging is essential for oral healthcare because it helps in the diagnosis and treatment of various dental conditions [1]. For example, dentists can recognize jaw-related conditions and identify anatomical characteristics such as teeth, maxillary sinus, and alveolar bone using panoramic dental X-ray images [2]. Furthermore, the precise measurements offered by this technique provide technical assistance in preoperative diagnosis, surgical planning, and postoperative evaluation [3,4].

Teeth image segmentation is a vital process in computer-assisted dentistry diagnostics and serves as an initial step in analyzing the tooth status. Dentists can use panoramic radiographs to assess a range of dental conditions, including missing teeth, dental development, impacted teeth, and adjacent relationships [2]. This is achieved through image segmentation. Current technology employs a ground-truth identification mechanism for panoramic X-ray images and utilizes a segmentation architecture to generate precise segmentation outcomes that can potentially facilitate clinical diagnosis. Panoramic dental X-ray scans and tooth image segmentation technology are essential components of computer-assisted dentistry diagnostics because they enable accurate measurements and provide a comprehensive view of the jaw and teeth. For example, accurate teeth segmentation from panoramic images is essential for diagnosing serious dental conditions like periodontitis, which is a severe gum infection leading to potential tooth loss. Through detailed segmentation, dentists can identify anomalies in the tooth and surrounding structures, such as enlarged periodontal ligament spaces or bone loss. However, while segmentation is crucial for initial evaluations, more specific imaging techniques, like bitewing X-rays or CBCT, are often required for a thorough diagnosis and treatment plan. As a result, teeth segmentation remains a cornerstone in dental diagnostic tools.

The precise categorization of teeth into distinct groups poses a significant challenge in dental image analysis despite its critical importance in various applications such as orthodontic treatment planning, dental implant surgery, and forensic odontology. Manual, semiautomatic, and automatic approaches have been devised to segment teeth in dental images [3]. Despite the progress made in this field, dental restorations, malocclusions, and pathological conditions can affect the segmentation performance.

Artificial intelligence applications in dentistry are growing, as they help practitioners increase patient safety while simplifying complicated procedures and offering predictable outcomes [5]. Medical image analysis uses deep learning techniques that provide several benefits, including anomaly detection, image segmentation, and classification [3]. Hence, AI systems can potentially improve health data outcomes, lower healthcare costs, and advance medical research [6]. For example, in dental image analysis, deep learning techniques have shown promising results in segmenting and classifying teeth and dental structures, resulting in enhanced diagnosis and treatment planning for various dental conditions [3,6].

Convolutional neural networks (CNNs) have emerged as the primary technique for image segmentation in dental imaging because of their ability to collect local spatial data and learn feature representation [7]. However, recent advancements in deep learning architectures, such as transformers, have demonstrated their potential to outperform CNNs in several computer vision tasks by effectively modeling long-range dependencies and global context [9]. The progress of CNN-based segmentation models in accurately segmenting teeth and other dental structures from background noise enables the precise analysis and diagnosis of dental conditions. With the increasing availability of large-scale dental image datasets and the ongoing advancements in deep learning techniques, CNN-based models are expected to play an even more significant role in dental imaging by facilitating rapid, accurate, and automated image analysis.

Boundary box filters, also known as region proposal methods, have been extensively used in medical image segmentation tasks to improve the performance of deep learning models by focusing on specific areas of interest. Boundary box filters were used to identify nodules on CT scans in [11, 14] when segmenting lung nodules accurately. Similarly, Oktay et al. [12] used boundary box filters to increase the pancreatic segmentation accuracy. Fan et al. [13] successfully segmented lesions in colonoscopy images using boundary box filters. By focusing on specific areas of interest in medical images, Nader et al. [10] demonstrated that boundary box techniques can enhance segmentation precision. In dental imaging, boundary box techniques have been used to segment teeth and other dental structures by identifying the regions of interest on panoramic radiography and dental cone-beam computed tomography (CBCT) [2]. These studies suggest that boundary box filters have the potential to considerably enhance the accuracy and efficiency of dental and medical image segmentation, thereby improving diagnosis and treatment planning for a variety of conditions.

Transformers have recently emerged as solid deep learning architectures with excellent results in various computer vision applications, including image segmentation [9]. Transformers successfully represent long-term dependencies and the global environment using self-attention processes, enabling them to record connections between pixels or features in an input image regardless of their geographical distance. This characteristic makes transformers a potentially suitable choice for dental image segmentation tasks where accurately capturing the context and relationships between teeth and their surrounding structures is crucial.

In this study, we present a cutting-edge deep neural network model designed to segment teeth into 32 distinct categories based on panoramic dental radiography images. Our proposed model achieves an accuracy rate of 97.26%, a Dice Similarity Coefficient of 0.9102, and a Jaccard Index of 0.8501, all of which represent significant improvements over previous methodologies, showing significant advancements in dental 2D X-ray image segmentation. These enhancements are the result of a novel methodology integrating CNNs and transformers, combined with TAB, as a novel addition. Utilizing these blocks facilitates the model's ability to focus on regions of interest, thereby effectively capturing both local and global contexts. TABs address significant challenges associated with dental image segmentation, such as overlapping structures and varied tooth shapes, thereby enhancing the overall performance and accuracy of the model.

Our proposed tooth segmentation model offers significant contributions to dental diagnostics and treatment planning. By providing precise tooth segmentation, our architecture enables early detection and diagnosis of various dental diseases. For instance, it can facilitate the detection of periodontal diseases or dental caries by identifying changes in tooth shape or the appearance of lesions. Moreover, the accuracy achieved in tooth segmentation can greatly assist in creating detailed treatment plans. Orthodontists, for example, can use the segmentation results to plan braces placement or determine the necessity of tooth extraction in overcrowded mouths. These applications underscore the clinical relevance and potential impact of our segmentation model in the field of dental medicine.

This paper is organized as follows. An overview of related works on dental picture segmentation is presented in Section 2, emphasizing deep learning-based methods. Section 3 describes the proposed deep learning architecture. The experimental setup, including the dataset, assessment criteria, and implementation information, is presented in Section 4. Section 5 presents experimental results and comparisons with other state-of-the-art models. Finally, Section 6 concludes the paper by addressing possible future research directions in this field.

## II. RELATED WORKS

Medical image segmentation has witnessed significant advancements with the advent of deep learning-based approaches. Yamanakkanavar and Lee [14] presented a novel

TABLE I
SUMMARY OF RELATED WORKS

| Method | Used for | Dataset | Advantages |
|---|---|---|---|
| Teeth U-Net[2] | Segmentation of teeth into binary categories using panoramic X-rays | Dental panoramic X-ray images | Uses a context semantics module and contrast enhancement module to improve accuracy |
| Region-based segmentation algorithm[6] | Segmentation and numbering of teeth in bitewing radiographs | Bitewing radiographs | Simple and efficient. Uses a region-based segmentation algorithm |
| Modified Unet [10] | Segmentation of multi-class teeth in panoramic X-rays | Panoramic X-rays | Uses U-Net architecture to improve accuracy |
| BB-Unet[11] | Segmentation of medical imaging | SegTHOR dataset and Cardiac dataset | Achieve good segmentation results |
| M-SegNet with global attention CNN model[14] | MRI segmentation | BraTS 2018 | More effective for brain MRI segmentation than previous methods |
| CE-Net[16] | Multi-organ Segmentation | MICCAI 2017 Multi-organ Segmentation Challenge dataset | More effective for medical image segmentation than previous methods |
| Piecewise training[17] | Semantic segmentation | PASCAL VOC 2012 | More efficient than traditional methods. More accurate than traditional methods. Robust to noise. |
| DeepLungSeg[18] | Lung segmentation | LUNA16 | Uses multiscale input, residual connection, and spatial attention mechanism to improve accuracy |
| Patch-wise U-Net[19] | Brain MRI segmentation | BraTS 2018 | Uses a patch-wise training approach, residual connection, and spatial attention mechanism to improve accuracy |
| U-Net[20] | Biomedical image segmentation | ISBI 2015 Challenge dataset | Uses contracting and expanding paths and skip connections to improve accuracy |
| Mask R-CNN[21] | Number of tooths and as well as the segmentation | | More effective for object detection and instance segmentation than previous methods. - Uses RPN, mask head, and RoIAlign layer to improve accuracy |
| ResNets[22] | Image recognition | ImageNet | More effective for image recognition than previous methods. - Uses residual learning to improve accuracy |
| Mask R-CNN[23] | Tooth detection and segmentation | Occlusal radiograph | Uses Mask R-CNN architecture to improve accuracy |
| Transfer learning[24] | Automated dental image analysis | Small dataset of dental images | Can be used on small datasets. Uses transfer learning to improve accuracy |
| Region growing algorithm and morphological operation[25] | Individual tooth segmentation from CT scan images | CT images of teeth in contact | Simple and efficient. Uses region growing algorithm and morphological operation to improve accuracy |
| U-NETS[26] | Segmentation of teeth in dental panoramic radiographs | Dental panoramic X-ray images | Uses a pre-trained U-Net model, data augmentation, and loss function specifically designed for teeth segmentation |
| EEDN[27] | Segmentation of maxillofacial images | Maxillofacial images | Uses a lightweight encoder, dilated convolution, and skip connection to improve efficiency and accuracy |
| TSASNet[28] | Segmentation of teeth in binary categories from dental panoramic X-ray images | Dental panoramic X-ray images | Uses a two-stage architecture and attention mechanism to improve accuracy |
| TSegNet[29] | Segmentation of teeth in 3D dental models | 3D dental models | Uses residual connection, spatial attention module, and dice loss function to improve accuracy and efficiency |

M-SegNet architecture that used a global attention CNN model for automated brain MRI segmentation. The base architecture, M-Net [41], is also used for brain image segmentation. Badrinarayanan *et al*. [15] developed SegNet, a deep convolutional encoder–decoder architecture, with significant success in medical imaging applications. Gu *et al*. [16] proposed CE-Net, a situation-encoding network for 2D medical-picture segmentation. Lin *et al*. [17] presented an efficient piecewise training approach for deep-structure models in semantic segmentation. Slic-Seg, a minimally interactive segmentation approach for the placenta using fetal MRI, was proposed by Wang *et al*. [18]. Lee *et al*. [19] proposed a patchwise U-Net structure for automated brain MRI segmentation. Deep learning-based techniques have demonstrated exceptional efficacy in various medical image-related assignments, underscoring their potential.

Dental image segmentation has recently emerged as a popular research area. In this regard, deep learning-based algorithms have exhibited encouraging results. This section provides a comprehensive review of the literature on dental image segmentation, focusing on deep learning-based techniques. The intricate nature and proximity to adjacent anatomical structures render dental image segmentation a formidable task. Nevertheless, deep learning methodologies have been demonstrated to overcome these obstacles and exhibit encouraging outcomes. Various deep learning architectures and techniques have been utilized in numerous studies on dental image segmentation. These include U-Net [20], Mask R-CNN [21], and ResNet [22]. The results of these studies suggest that deep learning-based methodologies can yield positive results in terms of precision and expediency in the dental segmentation of image tasks.

Researchers have employed several methods to enhance dental segmentation. Tekin *et al*. [23] segmented and numbered teeth in dental imaging panoramic images using a Mask R-CNN, yielding high-quality segmentation masks. Similarly, Yang *et al*. [24] developed an automated system for dental image analysis that included dental image diagnostic knowledge, drastically reducing the amount of human labor necessary for data preparation. Xia *et al*. [25] presented a method that successfully separated individual teeth from CT images of the upper and lower natural contact-scanned teeth.

CNNs have been extensively used in various medical image segmentation applications because of their ability to gather local spatial inputs and generate hierarchical representations [7]. Several CNN-based algorithms for tooth segmentation have been introduced for dental image analysis. Hou *et al*. [2] introduced Teeth U-Net, a segmentation approach for tooth panoramic X-ray images that uses a U-Net structure to capture contextual semantics and improve image contrast. Similarly, Tekin *et al*. [6] developed an improved tooth segmentation and numbering technique for bitewing radiographs using a machine-learning algorithm based on the U-Net architecture. These studies showed that CNNs efficiently segment teeth and dental structures using different dental images.

Recent advancements in deep learning architectures, such as transformers, have shown their potential to outperform CNNs in several computer vision tasks by effectively modeling long-range dependencies and global context [8]. Although Transformers have mainly been used for natural language processing applications, their use in medical-picture analysis is gaining popularity. Transformers have been used for image segmentation, classification, and anomaly detection tasks and have shown promising results in various medical domains. The utilization of bounding box techniques to concentrate on regions of interest has been observed in medical imaging. This approach serves to decrease the intricacy of segmentation tasks. Nader *et al*. [12] proposed an automatic tooth segmentation method for panoramic X-rays using deep neural networks with bounding boxes to enhance the accuracy of the segmentation process. El Jurdi *et al*. [11] presented BB-Unet, a U-Net design that includes bounding box priors to improve segmentation results for medical imaging tasks. The aforementioned studies demonstrated the capacity of bounding box methodologies to enhance the segmentation outcomes and augment the overall efficacy of deep-learning models.

U-Net [20] has been widely used for dental segmentation tasks. Koch *et al*. [26] employed a U-Net architecture to segment panoramic images of teeth, resulting in enhanced sample segmentation using a more compact and less complex network design. Similarly, Kong *et al*. [27] proposed an efficient encoder–decoder network (EED-Net) for the fast and accurate segmentation of maxillofacial images. Zhao *et al*. [28] developed a two-stage attention segmentation network (TSASNet) to locate and segment teeth in dental panoramic X-ray images that can combine pixel-level contextual information and identify fuzzy tooth areas. Cui *et al*. [29] proposed a tooth segmentation network (TsegNet) for 3D scanning of dental structures. Some researchers have also improved the U-Net architecture by enhancing the encoder and decoder, modifying the convolutional layers, and improving skip connections.

Attention techniques play a crucial role in boosting the performance of CNNs in medical image analysis tasks. These techniques allow the rescaling of extracted features through skip connections, thereby enhancing high-level representation learning. For example, Jin *et al*. [30] proposed residual attention U-Net (RAUNet) for liver tumor segmentation, which includes a backbone branch for learning original features and a soft mask branch to reduce noise and enhance positive features. Similarly, Liu *et al*. [31] introduced the deep residual attention network (DRANet), which improved the feature processing between the encoder and decoder, leading to more accurate lesion-type classification. Moreover, establishing extensive connections between encoders and decoders can enhance the links between different modules. To address this, Jose *et al*. [32] proposed the intervertebral disc network (IVD-Net), which utilized a dense technique to link encoders layer by layer, with each encoder processing a distinct image pattern. In addition, Zhang *et al*. [33] proposed a multiscale densely connected U-Net (MDU-Net) that fuses neighboring feature maps of multiple sizes at high and low levels to improve the encoder, decoder, skip connection performance, and segmentation accuracy. The related works explained above are mentioned in Table I for easy comparison.

By merging these related work sections, we can observe the evolution of dental image segmentation using deep learning techniques, from the early use of the Mask R-CNN to the more recent incorporation of transformers, bounding box techniques, and improvements to the U-Net structure. These advancements provide a strong foundation for our proposed method and exciting possibilities for future research in this field.

This diverse range of methods demonstrates the ongoing advancements and potential for further improvements in dental image segmentation using deep learning techniques. In this study, we introduced a novel deep learning architecture for tooth segmentation based on panoramic images to address the weaknesses of existing approaches. Our method goes beyond the integration of existing tools; it combines the advantages of CNNs and transformers with a unique tooth-bounding box technique that improves the accuracy of tooth segmentation while resolving the challenges that currently exist in dental image analysis.

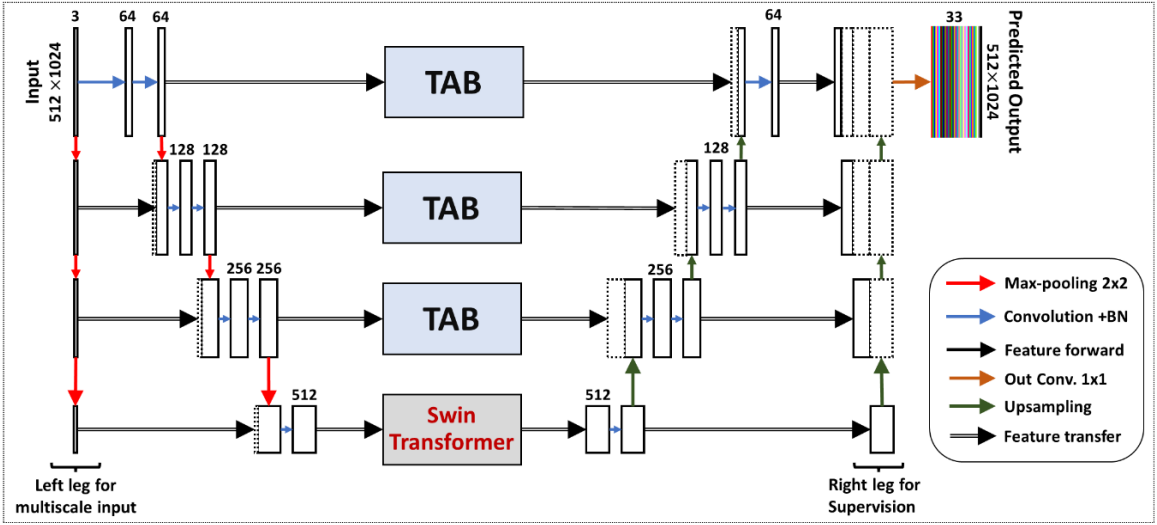

**FIGURE 1.** Overall architecture of the proposed teeth segmentation network. (TAB: Teeth Attention Blocks)

The critical contribution of our methodology is the strategic integration of CNNs, Transformers, and the novel TAB. CNNs are used to extract features from dental images and to capture specific local characteristics. When encapsulating the global context, a domain in which CNNs fall short, transformers are utilized. Additionally, our model incorporates novel TAB, allowing for a focused understanding of the overall dental arch structure, a factor in previous approaches.

Our significant contribution, TAB, is intended to improve segmentation outcomes. This technique addresses the sensitivity issues encountered in previous models by sharpening the focus on the teeth, thereby reducing the impact of noise and irrelevant regions.

Our approach makes several significant contributions to the field of dental image analysis.
1. This study proposes a cutting-edge multiclass teeth segmentation architecture that combines an M-Net-like structure with Swin Transformers. This architecture integrates various components to efficiently capture local and global contextual information, enabling the accurate delineation of teeth and their adjacent structures.
2. This study introduces an innovative component called TAB, which plays a crucial role in the proposed architecture. TAB enhances the segmentation performance by selectively attending to teeth-related features, further improving the accuracy of tooth segmentation.
3. This study incorporates a multiscale supervision strategy by utilizing the left and right legs of a U-Net structure. This strategy aids in precise feature representation and boosts segmentation performance by providing supervision at different scales.
4. Through a thorough examination, we demonstrated that our model outperforms state-of-the-art techniques in several important metrics.

### III. PROPOSED METHOD

#### A. OVERALL ARCHITECTURE
We propose a tooth segmentation architecture that integrates an M-Net-like structure with an encoder and decoder, Swin Transformer [34], and TAB to segment dental images accurately as shown in Figure 1. Our architecture aims to capture both local and global contextual information effectively, resulting in the precise delineation of teeth and surrounding structures. The U-Net-like structure consists of an encoder that extracts feature representations through downsampling and a decoder that reconstructs the segmentation mask through upsampling. The skip connections between the encoder and decoder layers preserve spatial information. Additionally, left- and right-leg supervision is employed for the encoder and decoder to facilitate accurate feature representation learning and improve segmentation performance.

Swin Transformer blocks, placed at the bottleneck, effectively capture long-range dependencies using the self-attention mechanism, which models nonlocal information

### B. PRE-PROCESSING
In our approach, pre-processing steps are performed to enhance the overall quality of panoramic images before training and testing the model. Figure 2 shows the block diagram of our pre-processing steps.

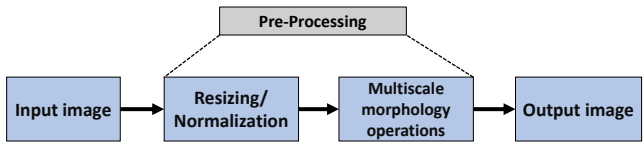

**FIGURE 2.** Block diagram of pre-processing

As shown in Figure 2, the pre-processing steps perform image resizing, normalization of pixel intensities to a range of [0, 255], and multiscale morphology in sequence. Multiscale morphology [38] employs a range of morphological operations at different image scales to reduce noise, improve contrast, and highlight the salient features of dental imagery. Such pre-processing is essentially required in refining the input data for the model, ensuring enhanced performance during the training and testing stages.

### C. ENCODER
Hierarchical dental image features are extracted in the encoder, where multiple convolutional layers are utilized with a 3×3 kernel size in each layer, which is the same size used in U-Net-like architectures. The encoder utilizes these layers to learn varying levels of features from the input dental images, from basic to complex. Batch normalization is performed after each convolution to improve the stability of the model and speed up learning. The activation function known as Rectified Linear Unit (ReLU) is responsible for introducing non-linearity into the model, enhancing its ability to learn intricate patterns. The inclusion of max-pooling layers is implemented to decrease the spatial dimensions of the feature maps and increase the receptive field. The features retrieved by the encoder substantially impact the overall performance of tooth segmentation. They are responsible for identifying different forms and patterns, hence facilitating accurate tooth segmentation.

### D. DECODER
The decoder recovers the spatial resolution of the feature maps and reconstructs the segmented teeth. This upsampling is achieved through transposed convolutions, effectively and relationships between distant regions in dental images. This enhanced the model's understanding of complex structures and relationships. Furthermore TAB in skip connections refine segmentation by focusing on object boundaries, leading to more precise delineations between different teeth and structures. This architecture effectively captures both local and global contextual information, resulting in accurate tooth segmentation.

increasing the height and width of the feature maps while preserving their depth. Simultaneously, we reconstruct segmentation masks from these up-sampled feature maps, resulting in pixel-wise class predictions for the input dental image.

Moreover, skip connections play a pivotal role in the decoder by bridging the gap between the encoder and decoder layers. These connections send the feature maps from the encoder to their corresponding decoder layers through the intermediate layers. This process allows for the incorporation of high-resolution details from the encoder's earlier layers with the abstract, lower-resolution features from the deeper layers. This fusion of features aids in the more accurate reconstruction of segmentation masks, as it captures both local details and global context, thereby enhancing the precision of the tooth.

### E. SWIN TRANSFORMER BLOCKS
Swin Transformer [34] is employed in deep learning architectures to effectively capture local and global contextual information using a self-attention mechanism. They divided the input feature maps into non-overlapping local windows, enabling efficient processing and utilizing multihead self-attention layers to learn multiple relationships simultaneously. Swin Transformers merge and shift windows after each self-attention layer to capture long-range dependencies, whereas position-wise feed-forward layers help learn complex nonlinear relationships The multihead self-attention layers in the Swin Transformer are followed by position-wise feed-forward layers and layer normalization, which allow the Swin Transformer to successfully manage the multiclass tooth segmentation task. Specifically, the multihead self-attention mechanism helps to capture intricate spatial relationships across different parts of the dental image, while the position-wise feed-forward layers enhance the local representations with non-overlapping local windows within input feature maps.

Following each self-attention phase, the Swin Transformer highlights its adaptability by merging and shifting windows, which is for capturing long-range dependencies. Additionally, the inclusion of position-wise feed-forward layers enhances the model's ability to identify complex nonlinear relationships. In aggregate, these methods enhance the effectiveness of the Swin Transformer in addressing dental image segmentation challenges. Since teeth exhibit diverse morphologies, it is essential to recognize subtle patterns and distant relationships in dental images. The Swin Transformers are specifically positioned to address the issues of tooth segmentation in our method.

The Swin Transformer blocks are strategically placed in the bottleneck of our design to effectively capture long-range dependencies. The utilization of nonlocal information management is of utmost importance, as it allows the model to effectively analyze complex connections between teeth and different dental diseases.

*F. TEETH ATTENTION BLOCK*

TABs, a key contribution of this study, act as boundary-aware or boundary refinement filters, playing a critical role in segmentation tasks to improve the recognition of boundaries between different objects or structures in an image. TABs enhance the focus of the tooth segmentation model on a dental image by observing a local receptive field around each pixel. The implementation of a filtering operation in this receptive field emphasizes the boundaries of the objects, specifically the edges of single teeth and their adjacent structures.

The attention mechanism employed by TAB plays a crucial role in enhancing the accuracy and precision of the segmentation masks. Specifically, this is achieved by effectively refining the differentiation between the individual and surrounding teeth. In simple terms, TAB reduces the effect of noise and meaningless information in dental images by selectively optimizing the features of teeth and their boundaries while minimizing irrelevant features or noise that are not beneficial to the teeth segmentation task. This improved attention helps achieve more precise and consistent segmentation results, providing a cleaner and clearer illustration of the individual teeth and their boundaries in dental panoramic images.

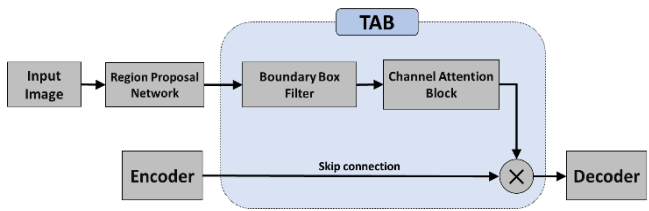

**FIGURE 3.** Teeth Attention Block in the proposed teeth segmentation network

The proposed TAB using a self-attention mechanism enables the model to assess the significance of various tooth parts within the image by leveraging acquired contextual knowledge. In the training stage, the TAB assigns increased attention scores to boundary pixels around separate teeth. This indicates that greater attention is given to the boundary regions while generating feature maps, hence improving the capability of the model to distinguish between individual teeth.

The improvement of the overall segmentation performance is facilitated by the attention capability of the TAB on these boundary regions. Specifically, it aids in enhancing the demarcation of tooth boundaries inside the segmentation masks, minimizing the occurrence of overlapping between neighboring teeth and improving the precision and accuracy of the segmentation process. The integration of the TAB method into our model yields a notable advantage, particularly in complex dental images characterized by densely arranged or slightly overlapping teeth, therefore mitigating the limitations of conventional segmentation techniques.

The TABs were incorporated into the skip connections in the tooth segmentation architecture between the encoder and decoder layers which can be seen in Figure 3. RPN detects boundary boxes that indicate potential regions containing dental structures. Then, the boundary boxes pass through the Channel Attention Block (CAB), fine-tuning the feature's attention. The utilization of specific filters in TABs significantly improves the segmentation accuracy of the model in accurately delineating complex object boundaries, particularly those pertaining to individual teeth and neighboring structures. The upgraded feature maps are subsequently multiplied with the features from the encoder via the skip connections. After the combination, the merged entities undergo processing by the decoder to achieve final segmentation.

TABs operate by considering a local neighboring receptive field around each pixel and implementing a filtering operation to highlight the boundaries of the teeth. The filtering operation of the TAB, which is one of our novel contributions, can utilize techniques such as convolutional layers and attention mechanisms to learn and target object boundaries. The incorporation of TAB within skip connections confers multiple benefits to the architecture of tooth segmentation, as follows:

- **Improved segmentation accuracy:** The model can distinguish between each tooth and its surrounding structures by concentrating on object boundaries, thereby producing more precise segmentation masks.
- **Smoother and sharper object boundaries:** The utilization of TAB has the potential to reduce the presence of unusual or uneven edges within the segmentation masks, thereby resulting in more refined and distinct object boundaries.
- **Better handling of overlapping or adjacent objects**: Teeth are frequently shown in proximity as well as overlapping in dental images. The implementation of TAB can enhance the model's ability to differentiate between teeth that are adjacent or overlapping with enhanced performance.

In summary, the proposed TAB plays a crucial role in enhancing the precision of tooth segmentation outcomes in the dental image segmentation framework. This is achieved by highlighting object boundaries and enhancing the differentiation between various teeth and structures.

*G. SUPERVISION AND LOSS FUNCTION*

The left and right legs of the U-Net structure were employed for supervision during the multiclass tooth segmentation task. Incorporating the multiscale supervision within both the downsampling (encoding) and upsampling (decoding) components of the U-Net, the model is supervised at multiple scales. Such a technique helps to capture and recognize the objects well and accurately segment the different classes of teeth found in dental images. This approach not only facilitates feature representation across various scales but also boosts the differentiation capacity of the model to distinguish between distinct teeth categories.

In this study, we train our model using a custom square Dice loss function. Dice loss, which is commonly employed in medical image segmentation tasks, calculates the overlap between the predicted and ground-truth images, making it ideal for addressing the class imbalance frequently observed in such tasks.

In the conventional Dice loss function, a score is calculated as twice the region of intersection between the predicted and true segmentation maps divided by the total number of pixels in both maps. The Dice loss was calculated as one minus the Dice score to obtain the best overlap between the predicted and actual segmentations.

Before formulating the loss, we changed the Dice loss function by squaring the pixel values. The following modification, described as square Dice loss, gives greater emphasis to every pixel, which makes the model more sensitive to segmentation boundary changes. The significance of precise segmentation boundaries in dental imaging cannot be overstated, because even minor deviations can significantly affect the quality of the resulting output.

The square Dice loss function includes a normalization factor that avoids division by a zero. The computation involves determining the intersection between the ground-truth segmentation map and the predicted segmentation map while considering the combined sum of the squares of both maps. This causes the neural network to prioritize accurate predictions for each pixel, which improves its precision and recall.

$$L = 1 - \left(2\sum(y_t \cdot y_p)^2 + \epsilon\right) / \left(2\sum(y_t^2 + y_p^2) + \epsilon\right) \quad (1)$$

where $L$ denotes the loss function. $y_t$ and $y_p$ represent the ground truth and predicted segmentation maps, respectively, ε is a smoothing factor for avoiding division by zero. The squared Dice loss is chosen for our method due to its efficiency, leading to superior segmentation results. In our investigation, we tried several loss functions, such as soft Dice loss [44], Tversky loss [43], and Log-Cosh Dice loss [43] functions, which are popularly used in medical segmentation tasks. It was observed that the squared Dice loss function achieved significantly better segmentation performance in DSC and JI, etc.

## IV. EXPERIMENTAL RESULTS

In this section, we discuss the experimental results for the proposed tooth segmentation architecture. The present study commenced by providing a detailed account of the dataset and the pre-processing procedures employed in the training and testing of the model. Subsequently, the evaluation metrics, experimental setup, and comparison with established methods are discussed. Finally, we evaluate the results while addressing the performance of the proposed model.

### A. DATASET

In this study, our dataset comprises dental panoramic images, a collaborative effort between a dental college and its students. These panoramic images were annotated meticulously using a supervisory platform, resulting in a detailed categorization of separate teeth across multiple classes. In total, our collection boasts 540 annotated images. To ensure computational efficiency and reduce memory demands, we resized these images to dimensions of 1024×512 pixels, taking care to preserve critical anatomical landmarks.

### B. EXPERIMENTAL SETUP

In this study, we utilized uniform settings to train and evaluate the proposed tooth segmentation network, ensuring a fair and consistent comparison with existing methodologies. The Keras framework and an NVIDIA GeForce RTX 3090 graphics processing unit (GPU) were utilized for model training and evaluation. The dataset was divided into training (70%), validation (15%), and testing (15%) datasets. We applied the Swin Transformer model with 2×2 regions trained over 50 epochs. The initial learning rate was set to $10^{-4}$ and subsequently adaptively decreased to $10^{-7}$ to address the potential overfitting issue. We maintained a batch size of two throughout the training process. Dropout layers are added after the encoder convolution layers to overcome overfitting. Furthermore, if the validation loss did not decrease over five consecutive epochs, we employed a strategy to decrease the learning rate by 10%.

The effectiveness of our proposed teeth segmentation model was quantitatively evaluated using a total of five standard evaluation metrics, which is Accuracy (ACC), Jaccard Index (JI) [35], Precision [36], Recall [36], and Specificity [37]. These metrics are defined as follows:

$$\text{Accuracy} = \frac{TP+TN}{TP+FP+FN+TN} \quad (2)$$

$$\text{JI} = \frac{|P \cap G|}{|P \cup G|} \quad (3)$$

$$\text{Precision} = \frac{TP}{TP+FP} \quad (4)$$

$$\text{Recall} = \frac{TP}{TP+FN} \quad (5)$$

$$\text{Specificity} = \frac{TN}{TN+FP} \quad (6)$$

where True Positive (TP) and True Negative (TN) represent the number of pixels accurately classified as teeth and non-teeth, respectively. Conversely, FP (False Positive) and False

Negative (FN) refer to the number of pixels incorrectly categorized as teeth and non-teeth, respectively. In the context of the Jaccard index, "P" typically represents the set of elements predicted by a model, while "G" represents the set of ground truth or actual values.

These metrics offer a comparable scale from zero to one, with one indicating an exact match between the predicted and actual values. Higher scores across these parameters denoted better segmentation performance, indicating that the model was efficient in accurately segmenting teeth and distinguishing between them and other structures in the dental images. By applying these evaluation metrics, we provide a comprehensive performance assessment of the proposed tooth segmentation model, facilitating its comparative analysis with other established methods in the field.

*C. RESULTS AND DISCUSSION*

This section presents a comprehensive analysis and discussion of the performance of the proposed model for dental segmentation. We compared our segmentation model to several well-established segmentation models, including the traditional U-Net [20], Attention U-Net [12], ResNet-50 Attention U-Net [39], Swin U-Net [40], and a Modified U-Net [10], which is identical to BB-Unet [11].The performance of the proposed model was evaluated using multiple critical metrics, including ACC, Dice Similarity Coefficient (DSC), JI, precision, recall, and specificity. It is noted that we aim to segment each tooth in an X-ray image into 32 categories based on the World Dental Federation (FDI) notation [42], where each tooth is categorized into #11 to #18, #21 to #28, #31 to #38 and #41 to #48. Since our model classifies each pixel into a specific number with multiclass 32-categorized pixels, we also measure the True Positives (TP), True Negatives (TN), False Positives (FP), and False Negatives (FN) to obtain objective evaluation metrics for segmentation. TP, TN, FP, and FN are computed for each tooth type, considering tooth number as one class and all other teeth as the other. This process was repeated for each tooth type in each class.

We performed a computational cost analysis, where we measured the training times for each method under identical experimental conditions. Our proposed method required approximately 60 minutes for training to achieve the segmentation performance in Table II. Other methods achieve DSC values of 0.7602, 0.7846, 0.7875, 0.6348, and 0.9004 and run-times of 45, 48-, 48-, 68-, and 64-minutes training times for [20], [12], [39], [40] and [10], respectively. These values are obtained from the above-mentioned experimental setup. Since the stopping criteria and learning rates are variable for each method during training, it is not difficult to judge the superiority of the complexity-performance trade-off. To investigate the change of segmentation performance for each method, we set a similar run-time by adjusting the stopping criteria and learning rate values. Table III shows the comparison of segmentation performances, such as DSC and JI, under almost identical run times. As shown in Table III, our proposed method can achieve significantly higher segmentation performance under similar complexity. It also indicates that our proposed method requires much lower run-times to achieve the same segmentation performance.

The accuracy score of our proposed (0.9726) is comparable to that of the other models. However, based on the Dice Coefficient (0.9102) and JI (0.8501), our model significantly outperformed the other models. These scores demonstrate that our model distinguishes true positives while minimizing false positives and false negatives. In addition, the precision and recall scores of 0.8046 and 0.9389, respectively, provided further evidence. This table demonstrates the superior performance of the proposed model compared to the others.

TABLE II
COMPARISONS BETWEEN THE PROPOSED METHOD AND CONVENTIONAL ONES

| MODELS | ACC | DSC | JI | PRECISION | RECALL | SPECIFICITY |
|---|---|---|---|---|---|---|
| U-NET [20] | 0.9720 | 0.7602 | 0.6871 | 0.7458 | 0.8366 | 0.9725 |
| ATTENTION_U-NET [12] | 0.9720 | 0.7846 | 0.7132 | 0.7557 | 0.8391 | 0.9725 |
| RESNET-50 ATTENTION U-NET [39] | 0.9721 | 0.7875 | 0.7172 | 0.7487 | 0.8544 | 0.9726 |
| SWIN U-NET [40] | 0.9712 | 0.6348 | 0.5296 | 0.6107 | 0.7192 | 0.9721 |
| MODIFIED-U-NET [10] | **0.9726** | 0.9004 | 0.8489 | 0.7898 | 0.9366 | 0.9728 |
| PROPOSED | **0.9726** | **0.9102** | **0.8501** | **0.8046** | **0.9389** | **0.9730** |

TABLE III
COMPARISONS OF RUNNING TIMES UNDER THE STOPPING CRITERIA IN EXPERIMENTAL SETUP

| MODELS | RUN-TIMES (IN MINUTES) | DSC | JI |
|---|---|---|---|
| U-NET [20] | 61 | 0.7556 | 06815 |
| ATTENTION_U-NET [12] | 59 | 0.7796 | 0.7071 |
| RESNET-50 ATTENTION U-NET [39] | 61 | 0.7804 | 0.7096 |
| SWIN U-NET [40] | 60 | 0.5432 | 0.4302 |
| MODIFIED-U-NET [10] | 61 | 0.9003 | 0.8490 |
| PROPOSED | 60 | 0.9102 | 0.8501 |

The performance of each model was critically analyzed. In this study, the traditional U-Net model exhibited commendable performance in terms of ACC, Dice score, and other metrics. However, it does not surpass the performance of the proposed model. Similarly, although the Attention U-Net and ResNet-50 Attention U-Net demonstrated improvements over the traditional U-Net model, they still did not match the performance of our proposed model. The performance of the Swin U-Net model did not align with those of the other models, whereas the Modified U-Net showed a performance close to that of our proposed model.

The superior performance of our proposed model can be attributed mainly to the incorporation of the Swin Transformer and boundary boxes and the application of a modified loss function utilizing the squared Dice loss. These design decisions enabled our model to learn and segment dental structures in the input images, thereby improving the performance across all evaluation metrics.

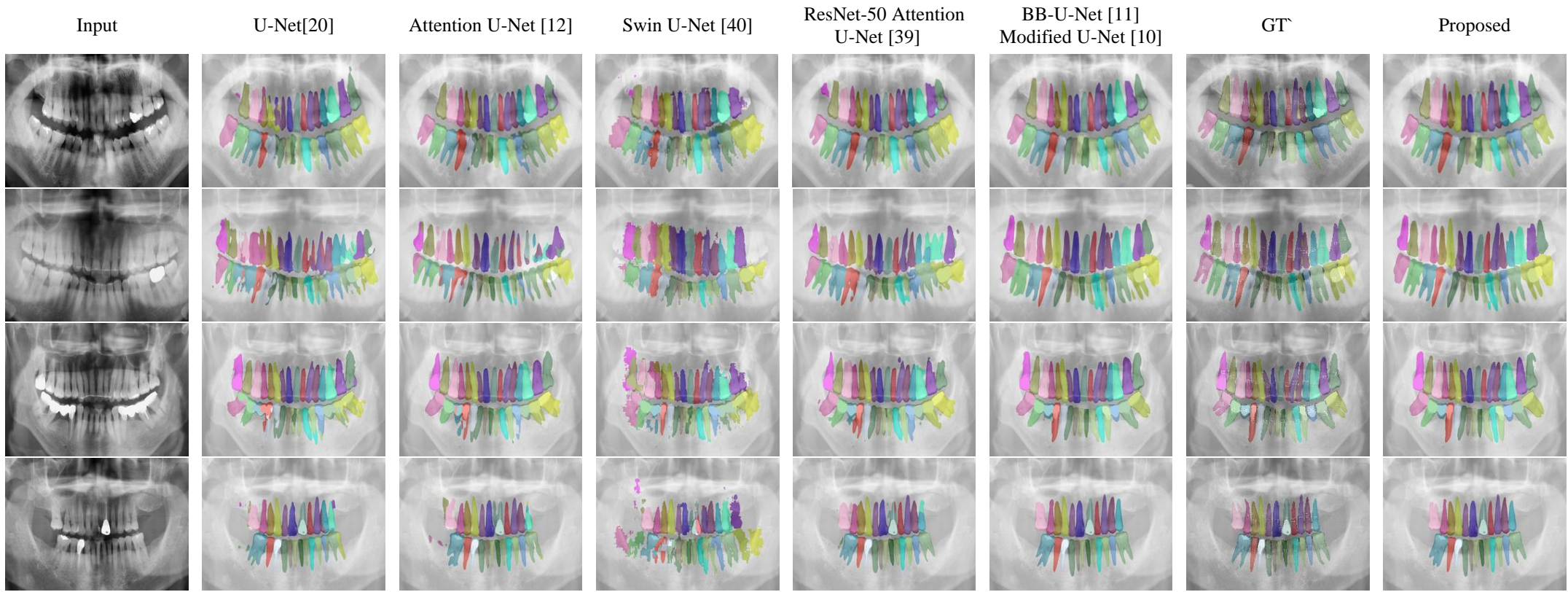

**FIGURE 4.** Visual Comparison between the proposed model and the conventional methods

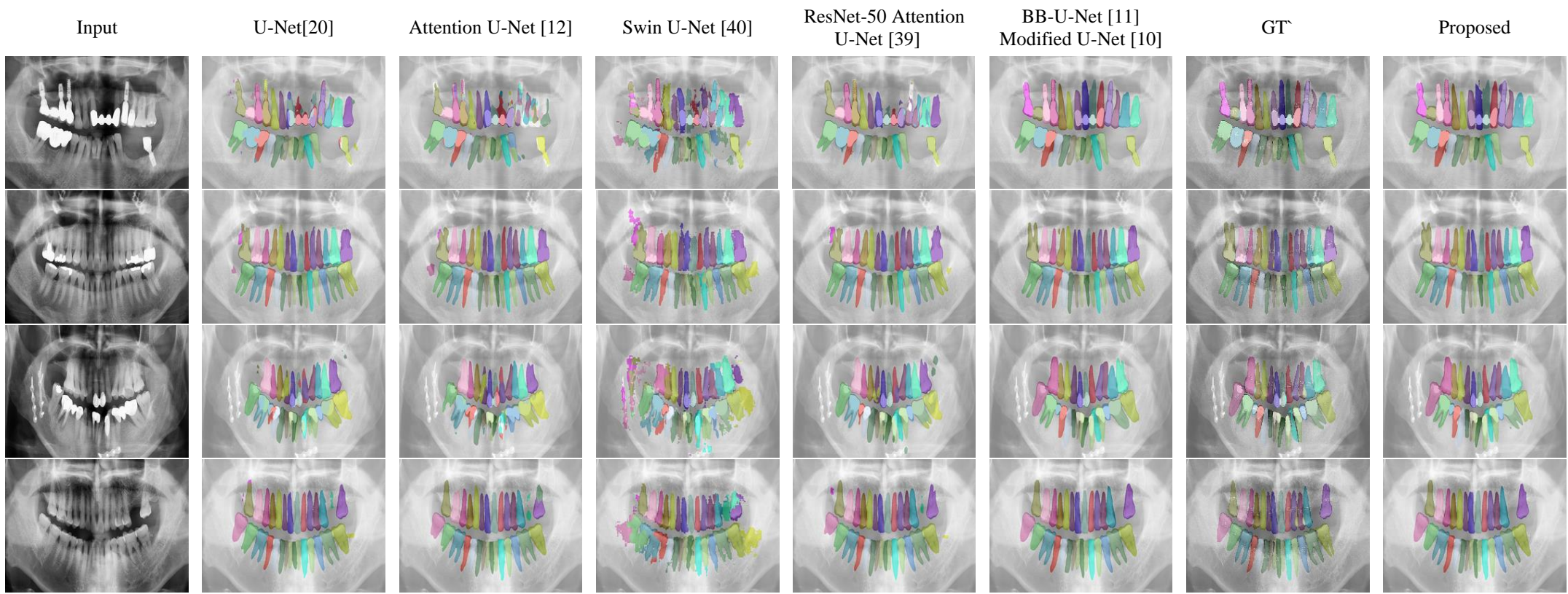

**FIGURE 5.** Visual Comparison between the proposed model and the conventional methods

In addition to a quantitative comparison, we visually compared the segmentation results of all models which can be seen in Figure 4 and Figure 5. This visual comparison provides evidence that the proposed model accurately segments dental structures. Furthermore, it consistently produced more accurate and consistent segmentation outcomes, highlighting the benefits of the Swin Transformer, boundary boxes, and modified loss function in our proposed dental segmentation model.

Furthermore, the improved generalizability of the approach used for different dental images can be primarily attributed to the addition of TAB and the squared Dice loss function. The use of TABs in dental imaging improves the accuracy of boundary delineation and increases the level of detail in individual tooth analyses. This innovative approach has demonstrated effectiveness in decreasing the impact of noise and artifacts that are commonly found in dental images. A closer analysis of row 1 of Fig. 3, illustrates the effectiveness of the proposed approach. For example, in the case of tooth #48 (FDI notation), most existing models struggle to precisely define the boundaries. However, our approach exhibits exceptional precision, providing segmentation outcomes that closely match the ground truth. This is largely due to the ability of TAB to focus on the local receptive fields surrounding each pixel, thereby highlighting the boundaries of objects and significantly improving the differentiation between individual teeth and their neighboring teeth.

This analysis highlights the potential advantages of our proposed approach over other approaches, particularly when dealing with complex dental structures and obtaining accurate and uniform segmentation results. The incorporation of TABs into the model demonstrated a marked improvement in the overall performance, indicating its potential as an asset in the progression of dental imaging analysis and diagnostics.

The results presented in the table and figures demonstrate that our custom dental segmentation model outperformed several state-of-the-art models in terms of evaluation metrics and visual comparisons. This significant boost in performance can be primarily attributed to the unique TABs. The use of TAB as boundary refinement filters significantly enhances the identification of each tooth and its adjacent structures. Although the Swin Transformer and the specific loss function play crucial roles, the use of TAB further propels the precision and efficiency of dental image analysis. This study provides a robust platform for future advances in dental image analysis and enhances the potential impact of dental procedures, including diagnosis, treatment planning, and patient monitoring.

*D. ABLATION STUDY*

An ablation study was performed to assess the contribution of each component to the proposed tooth segmentation network. Each variation in the network under similar training conditions successively included essential components of the basic U-Net model. The components investigated in this study include Deep Supervision, Swin Transformers, and TAB. We examined the effectiveness of each model, named Variations A to D, and the complete proposed network using several important metrics, such as ACC, DSC, JI, precision, recall, and specificity. The performance results for each variable are presented in Tables III and IV. We examined the effectiveness of each model, named Variations A to D, and the complete proposed network using several important metrics, such as ACC, DSC, JI, precision, recall, and specificity. The performance results for each variable are presented in Table V.

Table IV provides the details of the components of each variation. Variation A is the basic U-Net structure, and each variation includes an additional component, namely Deep Supervision (Variation B), Swin Transformers (Variation C), and TAB (Variation D). Table V shows the segmentation performances for the ablation study. As shown in Table V, the accuracy of the segmentation results gets higher over the variation number. Small gains were observed in DSC and JI in Variation B, where Deep Supervision is used. This improvement is due to better feature propagation throughout the network, enhancing the model's distinction between teeth classes. Swin Transformers yields a marginal enhancement in the DSC, as shown in the result of Variance C. It is due to the fact that Swin Transformers, with the self-attention mechanism, enable capturing local and global contextual information, which is a crucial factor for segmenting the complex structure of dental images where each tooth can influence the context of neighboring teeth. The improvement in the performance was notably observed in Variation D, where the proposed TAB is solely performed TAB enhances model performance by selectively focusing on teeth boundaries, enhancing the accuracy and precision of segmentation masks. TAB refines differentiation between teeth and surrounding structures by assigning higher attention scores to boundary pixels, resulting in more distinct edges of individual teeth. This enhances the model's overall performance, resulting in more accurate and detailed tooth segmentation results which can be observed from the performance metrics.

TABLE IV
COMPONENTS OF THE VARIATIONS IN THE ABLATION STUDY

| VARIATIONS | U-NET | DEEP SUPERVISION | SWIN TRANSFORMER | TAB |
|---|---|---|---|---|
| VARIATION A | ✔ | ✘ | ✘ | ✘ |
| VARIATION B | ✔ | ✔ | ✘ | ✘ |
| VARIATION C | ✔ | ✘ | ✔ | ✘ |
| VARIATION D | ✔ | ✘ | ✘ | ✔ |
| PROPOSED | ✔ | ✔ | ✔ | ✔ |

Table V shows the segmentation performances associated with each variation in the ablation study. The results demonstrate that each successive variant, with an additional component, results in a gradual increase in the performance metrics.

TABLE V
PERFORMANCE METRICS FOR VARIATIONS IN THE ABLATION STUDY

| VARIATIONS | ACCURACY | DSC | JI | PRECISION | RECALL | SPECIFICITY |
|---|---|---|---|---|---|---|
| VARIATION A | 0.9720 | 0.7602 | 0.6871 | 0.7458 | 0.8366 | 0.9725 |
| VARIATION B | 0.9722 | 0.7846 | 0.7132 | 0.7557 | 0.8391 | 0.9725 |
| VARIATION C | 0.9721 | 0.7644 | 0.6905 | 0.7569 | 0.8477 | 0.9726 |
| VARIATION D | 0.9725 | 0.9001 | 0.8476 | 0.7908 | 0.9312 | 0.9728 |
| PROPOSED | 0.9726 | 0.9102 | 0.8501 | 0.8046 | 0.9389 | 0.9730 |

We set the basic U-Net structure as a base in this study. As we added features like Deep Supervision and Swin Transformers, the performance of models shows improved results. Among the added featured tools, the most significant boost in performance is shown for the proposed TAB.

Although this paper proposes a novel tooth segmentation approach, it has certain limitations that guide our future works. The dental images used in this study contain complete teeth sets with relatively fewer images with dental diseases, which may restrict our learning capability of the model. Although our model achieves promising results in segmenting teeth into many classes, further studies can be feasible with a more extensive set of dental health issues.

## V. CONCLUSION

In this study, we introduce an innovative tooth segmentation model for dental panoramic images. It incorporates an M-Net-like structure with Deep Supervision, Swin Transformers, and TAB. The proposed model efficiently leverages local and global contextual information, resulting in significantly more accurate tooth segmentation. In particular, the proposed TABs show remarkable proficiency in highlighting complex dental anatomy and finely delineating tooth borders. The novel attention mechanism embedded in the TAB precisely highlights complex tooth structures, resulting in highly accurate segmentation outcomes. Using multiscale supervision and the squared Dice loss, our architecture effectively tackles class imbalances and enhances feature representation, ultimately achieving precise tooth delineation and surrounding structure definition. Our proposed method demonstrates its effectiveness and reliability in dental diagnosis applications on a real-world panoramic teeth X-ray dataset. Furthermore, our proposed method shows the feasibility of automated disease diagnosis and treatment planning owing the precise segmentation performance. For example, it enables the early detection of periodontal diseases or dental caries by identifying changes in tooth shape or the appearance of lesions. However, although our model achieves significantly better results over the state-of-the-art, the investigation of a more extensive set of dental health issues remains as further studies. The dental images used in this study contain complete teeth sets with relatively fewer images with dental diseases, which may restrict our learning capability of the model.

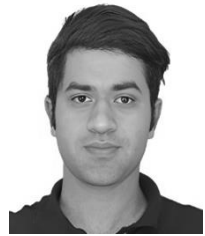

**AFNAN GHAFOOR** received the B.S. degree in electrical engineering from the University of Engineering and Technology, Lahore, Pakistan, in 2020. Since September 2021, he has been working as a Graduate Research Assistant in the Multimedia Image Processing Lab at Chosun University, South Korea, where he is also pursuing his M.S. degree in the Department of Information and Communications Engineering. His research interests include medical image processing, image classification, and image segmentation.

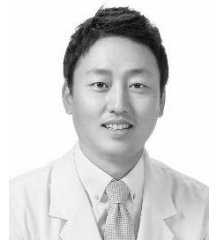

**SEONG-YONG MOON** received the B.S. and Ph.D. degrees in Dentistry from Chosun University and Chonnan University, respectively, He is currently a Professor in the Department of Oral and Maxillofacial Surgery, College of Dentistry, Chosun University. He is also the CEO of HT Core, a VR/AR-based dental simulation company.

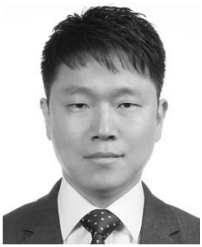

**BUMSHIK LEE** (Member, IEEE) received the B.S. degree in electrical engineering from Korea University, Seoul, South Korea, in 2000, and the M.S. and Ph.D. degrees in information and communications engineering from the Korea Advanced Institute of Science and Technology (KAIST), Daejeon, South Korea, in 2006 and 2012, respectively. He was a Research Professor at KAIST, in 2014, and a Postdoctoral Scholar at the University of California at San Diego, San Diego, CA, USA, from 2012 to 2013. He was a Principal Engineer at the Advanced Standard Research and Development Laboratory, LG Electronics, Seoul, from 2015 to 2016. In 2016, he joined the Department of Information and Communications Engineering, Chosun University, South Korea. His research interests include video processing, video security, and medical image processing.